\def\year{2020}\relax
\documentclass[letterpaper]{article} % DO NOT CHANGE THIS
\usepackage{aaai20}  % DO NOT CHANGE THIS
\usepackage{times}  % DO NOT CHANGE THIS
\usepackage{helvet} % DO NOT CHANGE THIS
\usepackage{courier}  % DO NOT CHANGE THIS
\usepackage[hyphens]{url}  % DO NOT CHANGE THIS
\usepackage{graphicx} % DO NOT CHANGE THIS
\def\UrlFont{\rm}  % DO NOT CHANGE THIS
\usepackage{graphicx}  % DO NOT CHANGE THIS
\usepackage{amsmath,amssymb}
\usepackage{xcolor}
\usepackage{subcaption}
\usepackage{multirow}
\usepackage{makecell}

\newcommand{\citet}[1]{{\citeauthor{#1} (\citeyear{#1})}}
\nocopyright

\title{Deconfounding age effects with fair representation learning when assessing dementia}
\author{Zining Zhu\textsuperscript{\rm 1,2}, Jekaterina Novikova\textsuperscript{\rm 1}, Frank Rudzicz\textsuperscript{\rm 3,2,4,5,1} \\
\textsuperscript{\rm 1}Winterlight Labs 
\textsuperscript{\rm 2}University of Toronto Computer Science \\
\textsuperscript{\rm 3}Li Ka Shing Knowledge Institute, St Michael's Hospital \\
\textsuperscript{\rm 4}Vector Institute for Artificial Intelligence
\textsuperscript{\rm 5}Surgical Safety Technologies Inc \\
{\tt \{zining, jekaterina\}@winterlightlabs.com, frank@cs.toronto.edu}
}

\begin{document}
\maketitle

\begin{abstract}
One of the most prevalent symptoms among the elderly population, dementia, can be detected by classifiers trained on linguistic features extracted from narrative transcripts. However, these linguistic features are impacted in a similar but different fashion by the normal aging process. Aging is therefore a confounding factor, whose effects have been hard for machine learning classifiers (especially deep neural network based models) to ignore. We show DNN models are capable of estimating ages based on linguistic features. Predicting dementia based on this aging bias could lead to potentially non-generalizable accuracies on clinical datasets, if not properly deconfounded.

In this paper, we propose to address this deconfounding problem with fair representation learning. We build neural network classifiers that learn low-dimensional representations reflecting the impacts of dementia yet discarding the effects of age. To evaluate these classifiers, we specify a model-agnostic score $\Delta_{eo}^{(N)}$ measuring how classifier results are deconfounded from age. Our best models compromise accuracy by only 2.56\% and 1.54\% on two clinical datasets compared to DNNs, and their $\Delta_{eo}^{(2)}$ scores are better than statistical (residulization and inverse probability weight) adjustments.  
\end{abstract}

\section{Introduction}
One in three seniors die of Alzheimer's and other types of dementia in the United States \cite{alzheimer2018}.
Although its causes are not yet fully understood, dementia impacts cognitive abilities in a detectable manner. This includes different syntactic distributions in narrative descriptions \cite{roark2007syntactic}, more pausing \cite{Singh2001}, higher levels of difficulty in recalling stories \cite{lunsford2015using}, and impaired memory generally \cite{lehr2012fully}.
Fortunately, linguistic features can be used to train classifiers to detect various cognitive impairments. For example, \citet{Fraser:2013} detected primary progressive aphasia with up to 100\% accuracy, and classified subtypes of primary progressive aphasia with up to 79\% accuracy on a set of 40 participants using lexical-syntactic and acoustic features. \citet{fraser15-JAD} classified dementia from control participants with 82\% accuracy on narrative speech. 

However, dementia is not the only factor causing such detectable changes in linguistic features of speech. Aging also impairs cognitive abilities \cite{harada2013normal}, but in subtly different ways from dementia. For example, aging inhibits fluid cognitive abilities (e.g., cognitive processing speed) much more than the consolidated abilities (e.g., those related to cumulative skills and memories) \cite{deary2009age}. In other words, the detected changes of linguistic features, including more pauses and decreased short-term memories, could attribute to just normal aging process instead of dementia. Unfortunately, due to the high correlation between dementia and aging, it can be difficult to disentangle symptoms are caused by dementia or aging \cite{murman2015impact}. Age is therefore a confounding factor in detecting dementia. In the next section, we will illustrate this confounding factor with an (undesirable) $P(\mathbf{X}, A)$ probability term.

The effects of confounding factors are hard for traditional machine learning algorithms to ignore, due to, e.g., sampling biases in the data. 
For example, word embeddings \cite{Caliskan183} showed that word embeddings capture human-like biases (towards age, gender, etc.) from training text corpus. As another example, many best-performing reading comprehension and textual inference models tend to rely on simple cues (e.g., the occurrence of token \texttt{not} or lexical overlapping) during classification \cite{mccoy2019Right,Niven2019Probing}, leading to potentially not generalizable accuracies. 
Similarly, as we will show in Experiments section, traditional neural network classifiers are capable of inferring age from linguistic features. Biasing on ages could introduce spurious accuracies on small dementia detection datasets. 

Dementia detection should be based on the input features \emph{and only the features themselves}, not any bias like age effects. This problem, traditionally formulated as confounding \cite{pearl2009causality}, unfortunately cannot be addressed well enough by classical deconfounding approaches (e.g., residualization, inverse probability weighting etc.), especially for clinical (e.g., dementia) datasets.

In this paper, we propose to address this deconfounding problem in a fair representation learning framework that protects age as a ``sensitive attribute''. A sensitive attribute (or ``protected attribute'') for fair machine learning can be race, age, or other variables whose impact should be ignored. For example, \cite{zemel2013fair} penalized classifiers for the differences in classification probabilities among different demographic groups, and \cite{Ashraf2018} encouraged the model to map data samples into latent representations with no information (i.e., max entropy) about protected attributions. In a fair representation learning framework, classifiers should be aware of cognitive impairments while actively filtering out any information related to aging. 
%After training, the classifiers produced better demographic similarities while compromising only a little overall accuracy.   

To enhance the abilities of the fair representation learning models, adversarial training can be incorporated. \cite{goodfellow2014generative} introduced generative adversarial networks, where a generator and a discriminator are iteratively optimized against each other. Several works \cite{Edwards2016,madras2018learning,Sattigeri2018} incorporated adversarial training to limit the classifiers' abilities to identify the sensitive attributes. 

However, previous approaches to fair representation learning involved either binary or categorical attributes. To apply to cognitive impairments detection, we want to represent age on a continuous scale (with some granularity if necessary). 
We formulate a fairness metric for evaluating the ability of a classifier to ignore a continuous-valued attribute. We also propose four models that compress high-dimensional feature vectors into low-dimensional representations which encrypt age from an adversary. We show empirically that our models achieve better fairness metrics than baseline deep neural network classifiers, while compromising accuracies by as little as $2.56\%$ and $1.54\%$ on our two datasets, respectively.

\section{Deconfounding}
\paragraph{Graphical Illustration}
The relationships between age, dementia, and linguistic features could be illustrated graphically. For example, if we assume that both age $A$ and dementia $D$ cause changes in a feature $X$ (and additionally assume that dementia is independent of age), their causal relationships form a v-structure \cite{koller2009pgm}: $A\rightarrow X\leftarrow D$. 

When multiple features are considered, we write them as a vector $\mathbf{X}$. Practically, the classifiers give estimations $\hat{D}$ such that the joint probability distribution $P(\hat{D}, \mathbf{X})$ approximates the true distribution $P(D, \mathbf{X})$. This appears appropriate if the dataset were not biased on $A$. Without $A$, the classifiers learn the correct distribution $P(\hat{D}, do(\mathbf{X}))$. However, the training data acquired from clinical trials are actually from:
\begin{align}
    \label{eq:train_data_joint_dist}
    P(D, \mathbf{X}, A) = \sum_{\mathbf{X}} \sum_A P(D, \mathbf{X}) P(\mathbf{X}, A)
\end{align}

Traditionally, there are several ways to eliminate the effects of the $P(\mathbf{X}, A)$ term: residualization (on either $\mathbf{X}$ or $D$), inverse probability weighting, and propensity score matching. As will be shown below and in Experiments, they are either inferior to our approach or not applicable to erasing the impacts of the continuous-valued confounder, age.

\paragraph{Inverse probability weighting} is an intuitive and popular approach \cite{Clare2018}. It assigns each data sample a weight of $P(A\,|\,\mathbf{X})$, so that the joint distribution (\ref{eq:train_data_joint_dist}) would become: (ideally unrelated to $A$) 
\[\sum_{\mathbf{X}} \sum_{A}\frac{P(D, \mathbf{X})P(\mathbf{X}, A)}{P(A|\mathbf{X})}=\sum_{\mathbf{X}}P(D, \mathbf{X}) P(\mathbf{X})\]

\paragraph{Residualization} Based on the potential outcomes framework \cite{Rubin2005}, residualization can be performed on the features $\mathbf{X}$. To residualize a feature $x$ against age $a$, one could fit a model $\hat{x}=f(a)$.\footnote{Practically, we implement $f(a)$ as either of linear regression $\hat{x}=\theta_0 + \theta_1 a$ or quadratic regression $\hat{x}=\theta_0 + \theta_1 a+ \theta_2 a^2$, where $\theta_i$ ($i \in \{0, 1, 2\}$) are parameters trained on control group samples.} $\hat{x}$ approximates the component of this feature $x$ brought by $a$. The residual, $x - \hat{x}$, is taken as the `clean' feature. In the Experiments section, we show that residualization is in general inferior to our approach in terms of performance. Moreover, residualization requires age input \emph{of test samples}, and the downstream traditional classifier still can infer age information from features. Ideally, we would like the model to take in linguistic features with no age inputs, and make decisions with no age bias.

Another possible approach, residualizing $D$, estimates the part of label caused by aging. However, this is not what our problem setting desires -- we try to detect dementia while ignoring the age impacts inherent in the linguistic features.

\paragraph{Propensity score matching} \cite{Rosenbaum1983} is another popular deconfounding approach estimating the impacts of treatments on outcomes. %The most common propensity score matching is done pair-wise. For example, when considering a binary treatment $T$, each sample in the treated group ($T=1$) is matched with a counterpart in the control group ($T=0$) with similar (ideally the same) propensity score\footnote{Defined here as the probability of treatment conditioning on the observed covariates $P(T=t\,|\,\mathbf{X})$, $d\in \{0, 1\}$.}. The difference between their outcome is an estimate of treatment effect given linguistic features $\mathbf{X}$. 
However, dementia $D$ is not a treatment, and the age $A$ is not binary. Moreover, propensity score matching require the dataset to satisify the ignorability assumption, but fair representation learning does not impose assumption on data distributions.

\section{Measuring disentanglement}
There are many measures of entanglement between classifier outcomes and specific variables. We briefly review some relevant metrics, and then propose ours.

\subsection{Traditional metrics}
\textbf{Correlation} (Pearson, Spearman, etc.) is often used to compare classification outputs with component input features. To the extent that these variables are stochastic, several information theoretic measures could be applied, including Kullback-Leibler divergence and Jensen-Shannon divergence. These can be useful to depict characteristics of two distributions when no further information about available data is given.

\textbf{Mutual information} can depict the extent of entanglement of two random variables. If we treat age ($A$) and dementia ($D$) as two random variables, then adopting the approach of \cite{Kwak2002} gives an estimation of mutual information $I(A, D)$. However, given the size of clinical datasets, it can be challenging to give precise estimates. 

An alternative approach is to assume the age variable $A$, dementia indicator variable $D$, and multi-dimensional linguistic feature $\mathbf{X}$ fit into some \emph{a priori} model (e.g., the v-structure mentioned above, $A\rightarrow\mathbf{X}\leftarrow D$), then the mutual information between $A$ and $D$ is:
\begin{align*}
I(A, D) &= \mathbb{E}_{p(A, D)} \text{ log } \frac{p(A, D)}{p(A)p(D)} \\
&= \mathbb{H}_A + \mathbb{H}_D + \mathbb{E}_{p(A, D)} \left[\text{ log } p(A, D) \right]
\end{align*}
where the entropy of age $\mathbb{H}_A$ and of cognitive impairment $\mathbb{H}_D$ remain constant with respect to the input data $X$, and $$\displaystyle p(A, D) = \sum_\mathbf{X} p(A, \mathbf{X}, D) = \sum_\mathbf{X} p(A\,|\,\mathbf{X}) p(D\,|\,\mathbf{X}) p(\mathbf{X})$$
However, this marginalized probability is difficult to approximate well, because (1) the accuracy of the term $p(A\,|\,\mathbf{X})$ relies on the ability of our model to infer age from features, and (2) it is hard to decide on a good prior distribution on linguistic features $p(\mathbf{X})$. We want to make the model age-agnostic, leading to a meaningless mutual information in the `ideal' case. 

In our frameworks, we do not assume specific probabilistic models correlating confounds and outcomes, and we propose more explainable metrics than the traditional statistical ones. 

\paragraph{Informativeness coefficient} \cite{Pryzant2018interpretable,Pryzant2018deconfounded} proposed a causal informativeness coefficient based on the potential outcome model \cite{Rubin2005}. Their coefficient measures how much information each representation of features $L(\mathbf{X})$ contains beyond the confounders:
$$\mathcal{I}(L) = \mathbb{E}[\text{Var}[ Y\,|\, L(\mathbf{X}), A] - \text{Var}[Y \,|\, L(\mathbf{X})]]$$

This coefficient measures the ability of encoder $L(.)$ filtering out the confounder information.

\subsection{Fairness metrics}
The literature in fairness representation learning offers several metrics for evaluating the extent of bias in classifiers. Generally, the fairer the classifier is, the less entangled the results are with respect to some protected features.

\begin{figure*}[h]
\begin{subfigure}[t]{.48\linewidth}
\centering
\includegraphics[width=\linewidth]{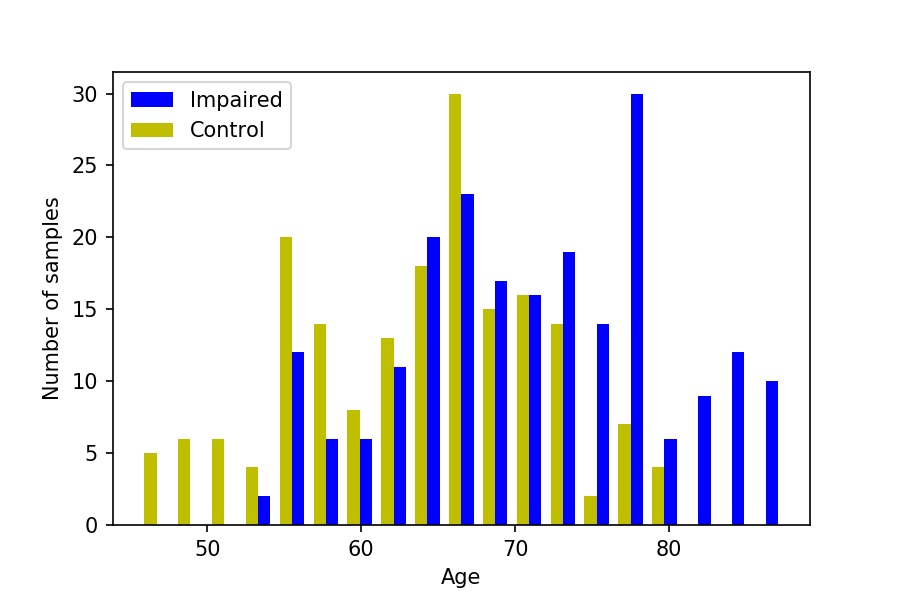}
\caption{DementiaBank
\label{subfig:dembank}}
\end{subfigure}
\begin{subfigure}[t]{.48\linewidth}
\centering
\includegraphics[width=\linewidth]{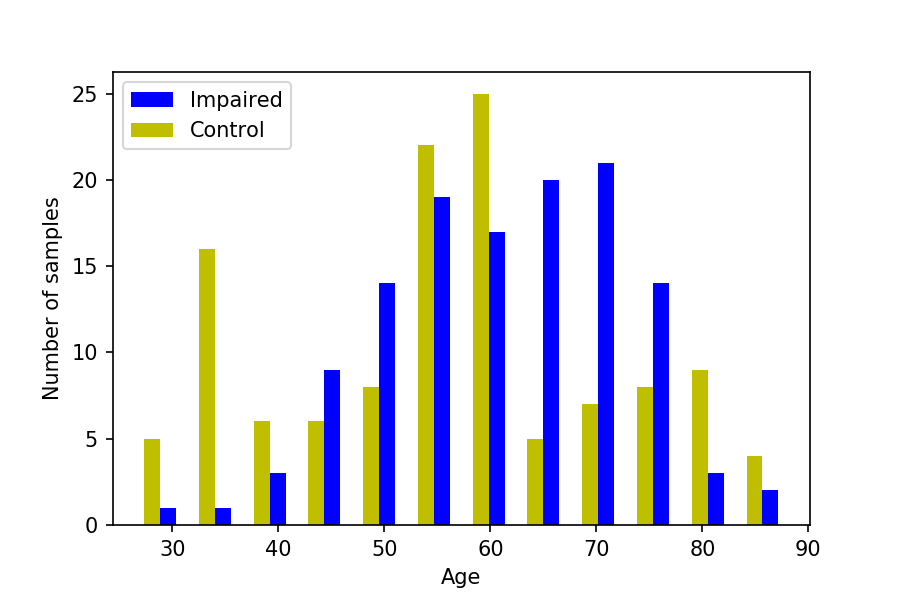}
\caption{Famous People
\label{subfig:Famous People}}
\end{subfigure}
\caption{Expository histogram plots for the ages of people in the impaired and control groups. Elderly participants in both datasets are more likely to have cognitive impairments.
\label{fig:age-eda-hist}}
\end{figure*}

\textbf{Demographic parity} \cite{zemel2013fair} stated that the fairest scenario is reached when the composition of the classifier outcome for the protected group is equal to that of the whole population. While generally useful, this does not apply to our scenario, in which there really \emph{are} more elderly people suffering from cognitive impairments than younger people (see Figure \ref{fig:age-eda-hist}). 

\textbf{Cross-entropy loss} \cite{Edwards2016} used the binary classification loss of an adversary that tried to predict sensitive data from latent representations, as a measure of fairness. This measure can only apply to those models containing an adversary component, not traditional classifiers. Moreover, this loss also depends on the ability of the adversary network. For example, a value of this loss could indicate confusing representations (so sensitive information are protected well), but it could also indicate a weak adversary.

\textbf{Equalized odds} \cite{hardt2016equality} proposed a method in which false positive rates should be equal across groups in the ideal case. 
\cite{madras2018learning} defined fairness distance as the absolute difference in false positive rates $p_a$ between two groups, plus that of the false negative rates $n_a$: 
\[
\Delta = \big|p_0 - p_1 \big| + \big|n_0 - n_1 \big|
\]
where a is the sensitive attribute $a=0$ ($a=1$).

\subsection{Our metric}
We propose an extension of the ``equalized odds'' ($eo$ for short) metric used by \cite{madras2018learning} to continuous sensitive attributes, suitable for evaluating an arbitrary two-class classifier. 

First, groups of age along a scale are divided so that each group has multiple participants with both positive and negative diagnoses, respectively. Let $a$ be the age group each participant is in.
Then, we want the expected false positive (FP) rates of the classifier be as constant as possible across age groups. This applies likewise to the false negative (FN) rates. We measure their variability with:
\[\label{eqn:eo_dist}
\Delta_{eo}^{(N_a)} = \sum_{a=1}^{N_a} \big| p_a - \hat{p} \big| + \sum_{a=1}^{N_a} \big| n_a - \hat{n} \big|,
\]
where $\hat{x}$ represents the mean of variable $x$.
Note that we do not average over the number of groups $N_a$, to emphasize the difference between e.g., $\frac{1}{2}\Delta_{eo}^{(2)}$ and $\frac{1}{4}\Delta_{eo}^{(4)}$. This is because given the size of many medical datasets, the divided group number $N_a$ significantly impacts this fairness metric. Therefore, it is important to provide $N_a$ while reporting this metric, especially on clinical datasets.

\subsection{Analysis of our metric}
\paragraph{Special cases} We illustrate our metric with several special cases here:
\begin{enumerate}
\item When there is only one age group, this metric is default to $\Delta_{eo}=0$.

\item When there are only two age groups, our metric equals that of \cite{madras2018learning}.

\item In the extreme case where there are as many age groups as there are sample points (assuming there are no two people with  identical ages but with different diagnoses), our metric becomes less informative, because the empirical expected false positive rates of that group is either $0$ or $1$. This is a limitation of our metric, and is the reason that we limit the number of age groups to accommodate the size of the training dataset.
\end{enumerate}

% Moved forward here -- display at top of next page
\begin{figure*} 
\begin{subfigure}{.32\textwidth}
\centering
\includegraphics[scale=0.5]{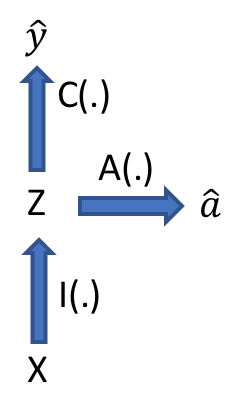}
\subcaption{age-indep-simple
\label{fig:age-indep-simple}}
\end{subfigure}
\begin{subfigure}{.33\textwidth}
\centering
\includegraphics[scale=0.45]{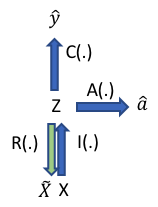}
\subcaption{age-indep-autoencoder and age-indep-entropy
\label{fig:age-indep-ae}}
\end{subfigure}
\begin{subfigure}{.33\textwidth}
\centering
\includegraphics[scale=0.5]{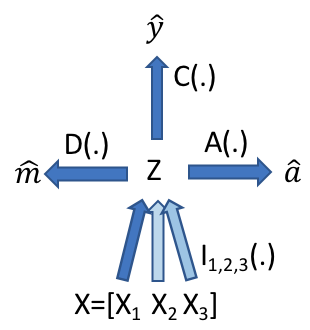}
\subcaption{age-indep-consensus-net
\label{fig:age-indep-cn}}
\end{subfigure}
\caption{Model structures. Each colored arrow denotes a neural network. The common components are interpreters $I(.)$, adversary $A(.)$, and classifier $C(.)$. In *-autoencoder and *-entropy (Figure \ref{fig:age-indep-ae}), a reconstructor $R(.)$ tries to reconstruct input data from the hidden representation. In *-consensus-nets (Figure \ref{fig:age-indep-cn}), a discriminator $D(.)$ tells apart from which modality the representation originates.
\label{fig:models}}
\end{figure*}

%\paragraph{Bounds} Our metric is bounded. The lower bound, $0$, is reached when all false positive rates are equal and when all false negative rates are equal across age groups. Letting $N_a$ be the number of age groups divided, an upper bound for $\Delta_{eo}^{(N_a)}$ is $N_a$ for any better-than-trivial binary classifier. The detailed proof is included in the Appendix \ref{sec:proof-bounds}.

\paragraph{Disentanglement} Our fairness metric illustrates deconfounding. A higher $\Delta_{eo}^{(N)}$ corresponds to a higher variation of incorrect predictions by the classifier across different age groups. Therefore, a lower value of $\Delta_{eo}^{(N)}$ is desired for classifiers biasing on age to a better extent. Throughout this paper, we use the terms `fair', `deconfounded', and 'disentangled' interchangeably.

%\paragraph{Design choices} We explain a few design choices here, namely linearity and indirect optimization.

%\emph{Linearity.} We encourage $\Delta_{eo}^{(N)}$ to be as linear as possible, for explainability of the fairness score itself. This eliminates possible scores consisting of higher order terms of FP / FN rates.

\paragraph{Indirect optimization.} We avoid directly optimizing the fairness score $\Delta_{eo}^{(N)}$. Although $\Delta_{eo}^{(N)}$ is correlated to the disentanglement between age and classification, it is based on FP / FN rates and hence bears their limitations -- FP / FN rates do not capture all aspects of classifiers. Instead of making the representations beneficial for $\Delta_{eo}^{(N)}$, we encourage the hidden representations to be age-agnostic (we will explain how to set up age agnostic models in the following section). As comparison, we also implement a model optimizing a differentiable version of EO distance (using probabilities instead of 0/1 predictions).

\section{Models}
In this section, we describe four different ways of building representation learning models, which we call age-indep-simple, age-indep-autoencoder, age-indep-consensus-net, and age-indep-entropy.

\subsection{age-indep-simple} 
The simplest model consists of an interpreter network $I(.)$ to compress high-dimensional input data, $\mathbf{x}$, to low-dimensional representations:
\[ \mathbf{z} = I(\mathbf{x})
\]
An adversary $A(.)$ tries to predict the \emph{exact} age from the representation:
\[ \hat{a} = A(\mathbf{z})
\]
A classifier $C(.)$ estimated the probability of label (diagnosis) based on the representation:
\[ P(\hat{y}) = \text{softmax} (C(\mathbf{z}))
\]

For optimization, we set up two losses: the classification negative log likelihood loss $\mathcal{L}_c$ and the adversarial (L2) loss $\mathcal{L}_a$, where:
\begin{align*}
\mathcal{L}_c = \mathbb{E}_x -\text{log} P(y) \hspace{3em} 
\mathcal{L}_a = \mathbb{E}_x || \hat{a} - a ||^2.
\end{align*}

We want to train the adversary to minimize the L2 loss, to train the interpreter to maximize it, and to train the classifier (and interpreter) to minimize classification loss. Overall,
\begin{align*} \min_{C, I} \mathcal{L}_c \text{ and } 
\max_{I} \min_{A} \mathcal{L}_a.
\end{align*} 
The training steps are taken iteratively, as in previous work \cite{goodfellow2014generative}.

\iffalse
%\begin{wrapfigure}[13]{r}{\textwidth}
%  \begin{minipage}{\dimexpr\linewidth-2\fboxrule-2\fboxsep}
  \begin{algorithm}[H]  % The [H] is required to fix position inside minipage
  \caption{Training age-indep-simple}\label{alg:age-indep-simple}
  \begin{algorithmic}[1]
  \State Initialize $I$, $A$, $C$
  \For {step := 1 to N} \Comment{N is a hyper-param}
  \For{minibatch $\mathbf{x}$ in training data $\mathcal{X}$}
  \State $\mathbf{z} = I(\mathbf{x})$, $a = A(\mathbf{z})$, $c = C(\mathbf{z})$
  \State Calculate $\mathcal{L}_a$, $\mathcal{L}_c$
  \State $\displaystyle \min_{I, C} \mathcal{L}_c - \mathcal{L}_a$  \Comment{backprop gradients}
  \For{ k:=1 to K}
  \State $\displaystyle \min_{A} \mathcal{L}_a$ \Comment{backprop gradients}
  \EndFor
  \EndFor
  \EndFor
  \end{algorithmic}
  \end{algorithm}
%\end{minipage}
%\end{wrapfigure}
\fi

\subsection{age-indep-autoencoder} The age-indep-autoencoder structure is similar to \cite{madras2018learning}, and can be seen as an extension from the age-indep-simple structure.
Similar to age-indep-simple, there is an interpreter $I(.)$, an adversary $A(.)$, and a classifier $C(.)$ network. The difference is that there is a reconstructor network $R(.)$ that attempts to recover input data from hidden representations:
$ \mathbf{\hat{x}} = R(\mathbf{z}) $.
The loss functions are set up as:
\begin{align*}
\mathcal{L}_c &= \mathbb{E}_x \text{ -log} P(y) \\
\mathcal{L}_a &= \mathbb{E}_x || \hat{a} - a ||^2 \\ 
\mathcal{L}_r &= \mathbb{E}_x || \mathbf{\hat{x}} - \mathbf{x} ||^2
\end{align*}
Overall, we want to train both the interpreter and the reconstructor to minimize the reconstruction loss term, in additional to all targets mentioned in the age-indep-simple network:
$ \min_{C,I,R} \mathcal{L} \text{ and } \max_I \min_A \mathcal{L}_a$ where $\mathcal{L} = \mathcal{L}_c + \mathcal{L}_r$.
%The detailed algorithm is similar to Algorithm \ref{alg:age-indep-simple} (see Appendix \ref{sec:optim-algorithms} for details).

\subsection{age-indep-consensus-net} 

This is another extension from the age-indep-simple structure, borrowing an idea from consensus networks \cite{CN2018}, i.e., that agreements between multiple modalities can result in representations beneficial for classification. By examining the performance of age-indep-consensus-net, we would like to see whether agreement between multiple modalities of data can be trained to be disentangled from age. 

Similar to age-indep-simple structures, there is also an adversary $A(.)$ and a classifier $C(.)$. The interpreter, however, is replaced with several interpreters $I_{1..M}$, each compressing a subset of the input data (``modality'') into a low-dimensional representation. The key of age-indep-consensus-network models is that these representations are encouraged to be indistinguishable. For simplicity, we randomly divide the input features into three modalities ($M=3$) with equal ($\pm$1) features.
A discriminator $D(.)$ tries to identify the modality from which the representation comes:
$ \hat{m} = D(\mathbf{z}) $.
The loss functions are set up as:
\begin{align*}
\mathcal{L}_c &= \mathbb{E}_x -\text{log} P(y) \\
\mathcal{L}_a &= \mathbb{E}_x || \hat{a} - a ||^2 \\ 
\mathcal{L}_d &= \mathbb{E}_x -\text{log} P(\hat{m})
\end{align*}

Overall, we iteratively optimize the networks:
\[\min_{C,I} \mathcal{L}_c \text{ and } 
\max_I \min_A \mathcal{L}_a \text{ and }
\max_I \min_D \mathcal{L}_d
\]

%The detailed algorithm is in the Appendix. 
Note that we do not combine the consensus network with the reconstructor because they do not work well with each other empirically. In one of the experiments by \cite{TCN2018}, each interpreter $I_m(.)$ is paired with a reconstructor $R_m(.)$ and the performance decreases dramatically. The reconstructor encourages hidden representations to retain the fidelity of data, while the consensus networks urges hidden representations to keep only the information common among modalities, which prohibits the reconstructor and consensus mechanism to function together.

\subsection{age-indep-entropy} The fourth model we apply to fair representation learning is motivated by categorical GANs \cite{springenberg2016categoricalGAN}, where information theoretic metrics characterizing the confidences of predictions can be optimized. This motivates an additional loss function term; i.e., we want to encourage the interpreter to increase the uncertainty (i.e., to minimize the entropy) while letting the adversary become more confident in predicting ages from representations.

Age-indep-entropy models have the same network structures as age-indep-autoencoder, except that instead of predicting the exact age, the adversary network outputs the probability of the sample age being larger than the mean:
\[ P(a|I,A,\mathbf{x}) = \text{softmax}(A(\mathbf{z}))
\]
This enables us to define the empirical entropy $\mathbb{H}[p]=\mathbb{E}_{x} p\text{log}\frac{1}{p}$, which describes the uncertainty of predicting age.

Formally, the loss functions are set up as follows:
\begin{align*}
\mathcal{L}_c &= \mathbb{E}_x \text{-log}P(y) \text{ and }
\mathcal{L}_r = \mathbb{E}_x || \mathbf{\hat{x}} - \mathbf{x} ||^2 \\
\mathcal{L}_a &= \mathbb{E}_x [\text{-log}P(a|I,A,\mathbf{x})] + \lambda_H \mathbb{H}[P(\hat{a}|I,A)]
\end{align*}
where $\lambda_H$ is a hyper-parameter. 
%For comparison, we also include two variants, namely the age-indep-entropy (binary) and age-indep-entropy (Honly) variants, each keeping only one of the two terms in $\mathcal{L}_a$. In our experiments, we show that these two terms in $\mathcal{L}_a$ are better applied together. 
Overall, the training procedure is similar to age-indep-autoencoder.
\[ \min_{C,I,R} \mathcal{L} \text{, and} \max_I \min_A \mathcal{L}_a \text{, where }
\mathcal{L} = \mathcal{L}_c + \mathcal{L}_r
\]

%\section{Implementation}
%All models are implemented in PyTorch \cite{pytorch}, optimized with Adam \cite{kingma2014adam} with initial learning rate of $3\times 10^{-4}$, and L2 weight decay $10$. For simplicity, we use fully connected networks with ReLU activations \cite{nair2010rectified} and batch normalization \cite{ioffe2015batch} before output layers, for all interpreter, adversary, classifier, and discriminator networks. Our frameworks can be applied to other types of networks in the future.

\section{Experiments}
\subsection{Datasets}

\paragraph{DementiaBank} DementiaBank\footnote{\url{https://dementia.talkbank.org/}} is a {\em relatively} large public dataset for assessing cognitive impairments using speech, containing 473 narrative picture descriptions from subjects aged between 45 and 90 \cite{dembank}. 
In each sample, a participant talks about what is happening in a clinically validated picture (i.e., the ``cookie-theft picture''\footnote{\url{http://languagelog.ldc.upenn.edu/myl/OldCookieTheft.png}}).
79 samples are excluded due to missing age information. 
In the remaining data samples, 182 are labeled `control' (negative), and 213 are labeled `dementia' (positive). Of all data samples containing age information, the mean is 68.26 and standard deviation is 9.00. 

\paragraph{Famous People} The Famous People dataset \cite{WLL-celebrity-dataset}
contains 252 transcripts from 17 people (8 with dementia including Gene Wilder, Ronald Reagan and Glen Campbell, and 9 healthy controls including Michael Bloomberg, Woody Allen, and Tara VanDerveer), collected and transcribed from publicly available speech data (e.g., press conferences, interviews, debates, talk shows). Seven data samples are discarded due to missing age information. %Additionally, data samples about Donald Trump is left out. 
Among the remaining samples, there are 121 labeled as control and 124 as impaired. Note that the data samples were gathered across a wide range of ages (mean 59.25, standard deviation 13.60). For those people diagnosed with dementia, there are data samples gathered both before and after the diagnosis, and all of which are labeled as `dementia'. %The Famous People dataset permits for early detection several years before diagnosis, which is a more challenging classification task than DementiaBank.  

\begin{table}[h]
\centering 
\begin{tabular}{l c c}
\hline
 & N. Samples (pos/neg) & Age \\ \hline 
DB & 213 / 182 & 68.26$\pm$9.00 \\
FP & 124 / 121 & 59.25$\pm$13.60 \\ \hline
\end{tabular}
\caption{Demographic information about the DementiaBank (DB) and Famous People (FP) datasets.
\label{tab:dataset-eda}}
\end{table}

% Moved forward here
\begin{table*}[t]
\centering
\begin{tabular}{l | l l l | l l l}
\Xhline{3\arrayrulewidth}
\multirow{2}{*}{Model} & \multicolumn{3}{c|}{DementiaBank} & \multicolumn{3}{c}{Famous People} \\
& Accuracy & $\Delta_{eo}^{(2)}$ & $\Delta_{eo}^{(5)}$ & Accuracy & $\Delta_{eo}^{(2)}$ & $\Delta_{eo}^{(5)}$ \\ \hline 
DNN baseline & .77$\pm$.05 & 0.17$\pm$0.14 & 0.94$\pm$0.22 & .65$\pm$.06 & 0.37$\pm$0.18 & 1.66$\pm$0.75\\ \hline 
*-simple & .75$\pm$.06 & \textbf{0.08$\pm$0.07} & \textbf{0.80$\pm$0.28} & \textbf{.64$\pm$.06} & 0.22$\pm$0.14 & 1.38$\pm$0.50\\
*-autoencoder & \textbf{.75$\pm$.05} & 0.11$\pm$0.08 & 0.88$\pm$0.24 & \textbf{.64$\pm$.07} & \textbf{0.21$\pm$0.16} & \textbf{1.27$\pm$0.47} \\
*-consensus-nets & .72$\pm$.05 & 0.12$\pm$0.08 & 0.90$\pm$0.55 & .62$\pm$.07 & 0.25$\pm$0.23 & 1.42$\pm$0.49 \\
*-entropy & .75$\pm$.04 & 0.13$\pm$0.10 & 0.97$\pm$0.61 & .62$\pm$.09 & 0.25$\pm$0.23 & 1.28$\pm$0.49 \\ 
%*-entropy (binary) & .72$\pm$.00 & 0.12$\pm$0.01 & 1.10$\pm$0.37 & .55$\pm$.07 & 0.26$\pm$1.53 & 1.41$\pm$0.40 \\
%*-entropy (Honly) & .74$\pm$.00 & 0.17$\pm$0.02 & 1.27$\pm$0.54 & .53$\pm$.06 & \textbf{0.20$\pm$0.16} & 1.39$\pm$0.49 \\
\hline 
\end{tabular}
\caption{Evaluation results of our representation learning models. The age-indep-simple and age-indep-autoencoder have better disentanglement scores while compromising the least accuracy. The ``age-indep'' prefix is replaced with ``*'' in model names.
\label{tab:our-models}}
\end{table*}

\begin{table*}[t]
\centering
\begin{tabular}{l | l l l | l l l}
\Xhline{3\arrayrulewidth}
\multirow{2}{*}{Deconfounding} & \multicolumn{3}{c|}{DementiaBank} & \multicolumn{3}{c}{Famous People} \\ 
& Accuracy & $\Delta_{eo}^{(2)}$ & $\Delta_{eo}^{(5)}$ & Accuracy & $\Delta_{eo}^{(2)}$ & $\Delta_{eo}^{(5)}$ \\ \hline 
Raw features & .77$\pm$.05 & 0.17$\pm$0.14 & 0.94$\pm$0.22 & .65$\pm$.06 & 0.37$\pm$0.18 & 1.66$\pm$0.75\\ \hline
Res-linear & .74$\pm$.03 & 0.21$\pm$0.16 & 1.08$\pm$0.38 & \textbf{.69$\pm$.04} & 0.27$\pm$0.19 & 1.72$\pm$0.74 \\
Res-quadratic & .74$\pm$.03 & 0.16$\pm$0.08 & 0.84$\pm$0.34 & .66$\pm$.07 & 0.32$\pm$0.17 & 1.49$\pm$0.57 \\
IPW-adjust & .70$\pm$.03 & 0.11$\pm$0.07 & \textbf{0.67$\pm$0.18} & .63$\pm$.08 & 0.32$\pm$0.15 & 1.87$\pm$0.49\\ \hline 
*-simple & .75$\pm$.06 & \textbf{0.08$\pm$0.07} & 0.80$\pm$0.28 & .64$\pm$.06 & 0.22$\pm$0.14 & 1.38$\pm$0.50\\
*-autoencoder & .75$\pm$.05 & 0.11$\pm$0.08 & 0.88$\pm$0.24 & .64$\pm$.07 & 0.21$\pm$0.16 & \textbf{1.27$\pm$0.47} \\
%SVM & .77$\pm$.05 & 0.17$\pm$0.13 & 0.93$\pm$0.29 & \textbf{.60$\pm$.04} & 0.23$\pm$0.19 & 1.28$\pm$0.29 \\
%Random Forest & .74$\pm$.03 & 0.19$\pm$0.14 & 1.07$\pm$0.36 & .56$\pm$.06 & 0.33$\pm$0.26 & 1.35$\pm$0.42 \\
%Adaboost & \textbf{.78$\pm$.07} & 0.14$\pm$0.11 & 0.96$\pm$0.22 & .54$\pm$.04 & 0.23$\pm$0.14 & 1.36$\pm$0.57\\
\hline 
Optimize-EO & .68$\pm$.03 & \textbf{0.08$\pm$0.04} & 0.91$\pm$0.27 & .64$\pm$.05 & \textbf{0.16$\pm$0.15} & 1.35$\pm$0.47\\
\hline 
\end{tabular}
\caption{Accuracy and fairness ($\Delta_{eo}^{(2)}$ and $\Delta_{eo}^{(5)}$) of several traditional deconfounding methods (Residualization on $\mathbf{X}$ with linear or quadratic model respectively, inverse probability weighting) versus our two best models (age-indep-simple and age-indep-autoencoder).
\label{tab:deconfounding}}
\end{table*}

\paragraph{Feature extraction}
We extract 413 linguistic features from the narrative descriptions and their transcripts. These features were previously identified as the most useful for this task \cite{roark2007syntactic,fraser15-JAD,lunsford2015using}.
Several examples of these features include:
\begin{itemize}
    \item Acoustic: pause-word ratio, speech rate, MFCC statistics.
    \item Syntactic: Yngve depth statistics, the occurrence of various context-free grammar.
    \item Semantic: vocabulary richness (Honor{\'e}'s statistics and Brun{\'e}t's index), part-of-speech derived features.
\end{itemize}
All feature values are $z$-score normalized. We refer to them as ``raw'' features in the following paragraphs.

\subsection{DNNs can bias on inferred ages}
\label{sec:detect-age}
As part of expository data analysis, we show that these linguistic features contain information indicating age.
Simple fully connected neural networks can predict age with mean absolute error of $15.5\pm 1.3$ years (on DementiaBank\footnote{Hidden layer sizes 64, 32, 8. 5-fold cross validation.}) and $14.3\pm 2.5$ years (on the Famous People dataset\footnote{Hidden layer sizes 32, 20, 2. 5-fold cross validation}).
This indicates that even simple neural networks are able to infer information about age from linguistic features. Neural classifiers can therefore also easily bias on age, given the utility of age in downstream tasks.

Why is it necessary to prevent this bias? The reason is, as illustrated in our Deconfounding section, that the feature-age joint distribution in clinical dataset $P(\mathbf{X}, A)$ suffers from sampling bias and deviate from the real-world distribution. A DNN model could directly classify whomever it considers to be more than 80 years old as ``dementia'', to improve accuracy on DementiaBank (similarly, for those between 65 and 73 years old on the Famous People dataset)\footnote{More specifically, $93.6\pm0.0\%$ elder-than-80 seniors on DB, and $80.9\pm0.0\%$ of the 65-to-73 seniors on FP are classified as positive, 5-fold cross validation, ten runs.}. However, this accuracy is obviously not generalizable -- indeed, much fewer real-world seniors are cognitively impaired.

\subsection{Setting up experiments against benchmarks}
We evaluate the performances of our four proposed neural network models against following benchmarks:
\begin{enumerate}
    \item DNN baseline using the features we extracted.
    \item Statistical adjustment methods: residualization according to ages using linear and quadratic models (Res-linear and Res-quadratic), inverse probability weighting (IPW-adjust).
    \item An approximate fairness lower-bound, directly optimizing the $\Delta_{eo}^{(2)}$ score (optimize-EO). The $\Delta_{eo}^{(2)}$ score is itself not differentiable, so we computed the probability instead of binary predictions when calculating the loss function, following the approach of \cite{zemel2013fair}.
\end{enumerate} 

For DNN baseline, we use a small multiple layer perceptrons (MLP)\footnote{One hidden layer with 5 neurons.}.
For statistical adjustment methods, we train MLPs with the same configurations using data processed with several traditional deconfounding methods.
For our four models, we train on raw features.
All MLP models are implemented with with sklearn \cite{scikit-learn}, while the remaining models use PyTorch \cite{pytorch}, optimized with Adam \cite{kingma2014adam}.

The performances are evaluated by both accuracy and our fairness metrics ($\Delta_{eo}^{(2)}$ and $\Delta_{eo}^{(5)}$).\footnote{These correspond to dividing ages into $N=2$ and $N=5$ groups respectively. $N=2$ and $N=5$ are arbitrary choices.} The results are listed in Table \ref{tab:deconfounding}. All accuracy and fairness results in this paper are based on 5-fold cross validation. No speech samples from the same person occur in both train and test sets.

\subsection{Performance and discussion}
\paragraph{Comparing to DNN baseline} The evaluation results for our models against DNN baseline are shown in table \ref{tab:our-models}. Our fair representation learning models compromise accuracy, in comparison to DNN baselines. This confirms that part of the classification accuracy of DNNs come from biasing with regards to age. 
On DementiaBank, the age-indep-autoencoder reduces accuracy the least (only 2.56\% in comparison to the DNN baseline). On the Famous People data, age-indep-simple and age-indep-autoencoder models compromise accuracies by only 1.54\% respectively, which are not statistically different from the DNN baseline\footnote{$p=20, 0.50$ on 38-DoF one-tailed $t$-tests, respectively.}. 

Why do the simpler models (i.e., *-simple and *-AE) outperform those more complicated ones (i.e., *-CN and *-entropy)? This might because the additional structural parameters bring in bias, which somewhat draws back the disentanglement.

\paragraph{Our models improve deconfounding scores}
Our fair representation learning models also improve the deconfounding scores\footnote{On DementiaBank, $p=0.004, 0.013$ and $0.020$ for *-simple, *-entropy and *-CN on $\Delta_{eo}^{(2)}$ respectively; these are significant. $p=0.06$ on age-indep-entropy on $\Delta_{eo}^{(2)}$; this is marginally significant. However, these differences are not as significant on $\Delta_{eo}^{(5)}$ (0.01, 0.40, 0.30, and 0.42.). On Famous People data, the $p$ values for our four models are all $\leq 0.01$ on $\Delta_{eo}^{(2)}$ and $\leq 0.05$ on $\Delta_{eo}^{(5)}$. These are all 38-DoF one-tailed $t$-tests.}, the improvements are mostly significant when measured by the two-group scores $\Delta_{eo}^{(2)}$. 
Also, the five-group scores $\Delta_{eo}^{(5)}$ are less stable for both datasets. Following is a possible explanation.
DementiaBank has $\sim$400 data samples and the Famous People dataset contains $\sim$250. In 5-fold cross validation, each of the five age groups has only $\sim$16 samples during evaluation. When the number of groups, $N$ is kept small (e.g., $\sim$100 samples per label per group, as in DementiaBank $N=2$), the fairness metrics are stable. 

\paragraph{Comparing to statistical adjustments}
Our best model is better than statistical adjustments (except $\Delta_{eo}^{(5)}$ than inverse probability weighting (IPW) adjustment on DB and accuracy than linear residualization on FP). Another advantage of our models is that fair representation learning models do \emph{not} require age inputs of testing samples, where all residualization adjustments require the input of the protected attribute. IPW adjustment does not need age inputs, but compromises accuracies more than others.

\paragraph{Comparing to direct optimization}
A noteworthy result is that our models has comparable fairness scores in comparison to the approximate lower-bound method, optimize-EO. The $\Delta_{eo}^{(2)}$ and  $\Delta_{eo}^{(5)}$ scores are not statistically different\footnote{When N=2, $p>0.06$ on DB, $p>0.05$ on FP. When N=5, $p>0.07$ on DB, $p>0.52$ on FP, for all four models.}. However, our models either have better accuracies than (on DB, $p<0.001$ for all four models, 1-tailed 38 DoF t-tests) or have comparable amount to (on FP, $p=0.41, 0.41, 0.11, 0.21$, 1-tailed 38 DoF t-tests on *-simple, *-AE, *-CN, *-entropy models respectively) optimize-EO.  
%Note that optimize-EO minimizes a differentiable variant of fairness score (because $\Delta_{eo}^{(N)}$ are not differentiable), hence its results only approximate the lower bounds of $\Delta_{eo}^{(N)}$

%\paragraph{Side notes}
%The model age-indep-entropy is best used with a loss function containing both the binary classification term and the uncertainty minimization term. As shown in Table \ref{tab:our-models}, although having similar fairness metrics\footnote{On DementiaBank, $p=0.19, 0.06$ for $\Delta_{eo}^{(2)}$ and $\Delta_{eo}^{(5)}$ of age-indep-Honly against age-indep-entropy, $p=0.24, 0.22$ for age-indep-binary. On Famous People, $p=0.24, 0.39$ for age-indep-Honly, and $p=0.33, 0.32$ for age-indep-binary. None of them are significant on 38-DoF one-tailed $t$-tests.}, the two variants with only one term could have lower accuracy than age-indep-entropy.

%\paragraph{Best deconfounders} In general, age-indep-simple and age-indep-autoencoder achieve the best fairness metrics. Noticeably, the better of them surpass traditional classifiers in both $\Delta_{eo}^{(2)}$ and $\Delta_{eo}^{(5)}$.

\section{Conclusion}
We identify the problem of age being confounded in the detection of cognitive impairments. DNN classifiers are able to estimate age from linguistic features, and could bias on them to detect dementia. To address this deconfounding problem, we formulate it in a fair representation learning setting, and propose a fairness score to measure the extent of deconfounding. 

We put forward four fair representation learning models that learn low-dimensional representations of data samples containing as little age information as possible. Our best models compromise as little as 2.56\% accuracy (on the DB dataset) and 1.54\% accuracy (on the FP dataset). Moreover, they have better $\Delta_{eo}^{(2)}$ scores than statistical adjustment methods. Their deconfounding scores are comparable to optimize-EO, a method approximating the fairness lower bound, but the accuracies of our best models are comparable (on FP) or significantly higher (on DB).

Our methods show a possibility of making neural network models more generalizable by overcoming sampling bias in limited, expensive clinical datasets.

\newpage
\bibliographystyle{aaai}
\bibliography{bibliography}

\end{document}